\documentclass[10pt,twocolumn,letterpaper]{article}

\usepackage{ijcb}
\usepackage{times}
\usepackage{epsfig}
\usepackage{graphicx}
\usepackage{amsmath}
\usepackage{amssymb}
\usepackage{array}
\usepackage{bm}
\usepackage[caption=false,font=footnotesize]{subfig}

\usepackage{url}

\usepackage{multirow}
\usepackage{booktabs}
\usepackage{color}
\usepackage{pifont}
\usepackage{changepage}

\def\hqwmca{\textit{HQ-WMCA }}
\def\siwm{\textit{SiW-M }}

\usepackage{xspace}
\usepackage{booktabs}
\usepackage{balance}
\usepackage{paralist}
\usepackage{multirow}
\usepackage{amsmath}
\usepackage[table]{xcolor}

\ijcbfinalcopy 

\ifijcbfinal\fi

\begin{document}

\title{On the Effectiveness of Vision Transformers for Zero-shot Face Anti-Spoofing} 

\author{Anjith George and S\'ebastien Marcel \\
Idiap Research Institute \\
Rue Marconi 19, CH - 1920, Martigny, Switzerland \\
{\tt\small  \{anjith.george, sebastien.marcel\}@idiap.ch  }
}

\maketitle
\thispagestyle{empty}

\begin{abstract}
The vulnerability of face recognition systems to presentation attacks has limited their application in security-critical scenarios. Automatic methods of detecting such malicious attempts are essential for the safe use of facial recognition technology. Although various methods have been suggested for detecting such attacks, most of them over-fit the training set and fail in generalizing to unseen attacks and environments. In this work, we use transfer learning from the vision transformer model for the zero-shot anti-spoofing task. The effectiveness of the proposed approach is demonstrated through experiments in publicly available datasets. The proposed approach outperforms the state-of-the-art methods in the zero-shot protocols in the \textit{HQ-WMCA} and \textit{SiW-M} datasets by a large margin. Besides, the model achieves a significant boost in cross-database performance as well.
\end{abstract}

\section{Introduction}

Face recognition offers a simple yet convenient way for access control. Though face recognition systems have become ubiquitous \cite{jain2011handbook}, its vulnerability to presentation attacks (a.k.a spoofing attacks) \cite{handbook2}, \cite{ISO}  limits the application of these systems in safety-critical applications. An unprotected face recognition (FR) system might be fooled by merely presenting artifacts like a photograph or video in front of the camera. The artifact used for such an attack is known as a presentation attack instrument (PAI).

As the name indicates, presentation attack detection (PAD) systems try to protect FR systems against such malicious attempts. Though a wide variety of presentation attacks are possible, the majority of the research efforts have focussed on the detection of 2D attacks such as prints and replays, mainly due to the easiness of producing such attack instruments. Most of the research in PAD focus on the detection of these attacks using the RGB spectrum alone, using either feature-based methods or Convolutional Neural Network (CNN) based approaches. Several feature-based methods using color, texture, motion, liveliness cues, histogram features \cite{boulkenafet2015face}, local binary pattern \cite{maatta2011face}, \cite{chingovska2012effectiveness} and motion patterns \cite{anjos2011counter} have seen proposed over the years for performing PAD. However, most of the recent state-of-the-art results are from CNN-based methods. Specifically, CNNs using auxiliary information in the form of binary or depth supervision have shown to improve performance greatly \cite{atoum2017face,george-icb-2019}. Nevertheless, the majority of these methods perform well only in the case of 2D attacks and the performance of such methods degrade when evaluated against sophisticated 3D and partial attacks \cite{Liu_2019_CVPR}. Even in the case of 2D attacks, these models often fail to generalize towards unseen attacks and environments.

Recently, several multi-channel approaches have been proposed to address the limitations of PAD systems \cite{george_mccnn_tifs2019, nikisins2019domain,george2020cmfl}. Though such methods achieve superior performance as compared to RGB methods, the cost of additional hardware required limits the application of such methods to protect legacy RGB-based FR systems. Hence, it is desirable to have a robust RGB-based PAD method that is robust against a wide variety of 2D, 3D, and partial attacks. Ideally, a PAD system should be able to generalize well to unseen attacks and environments.

\begin{figure}[t]
\centering
        \includegraphics[width=0.98\linewidth]{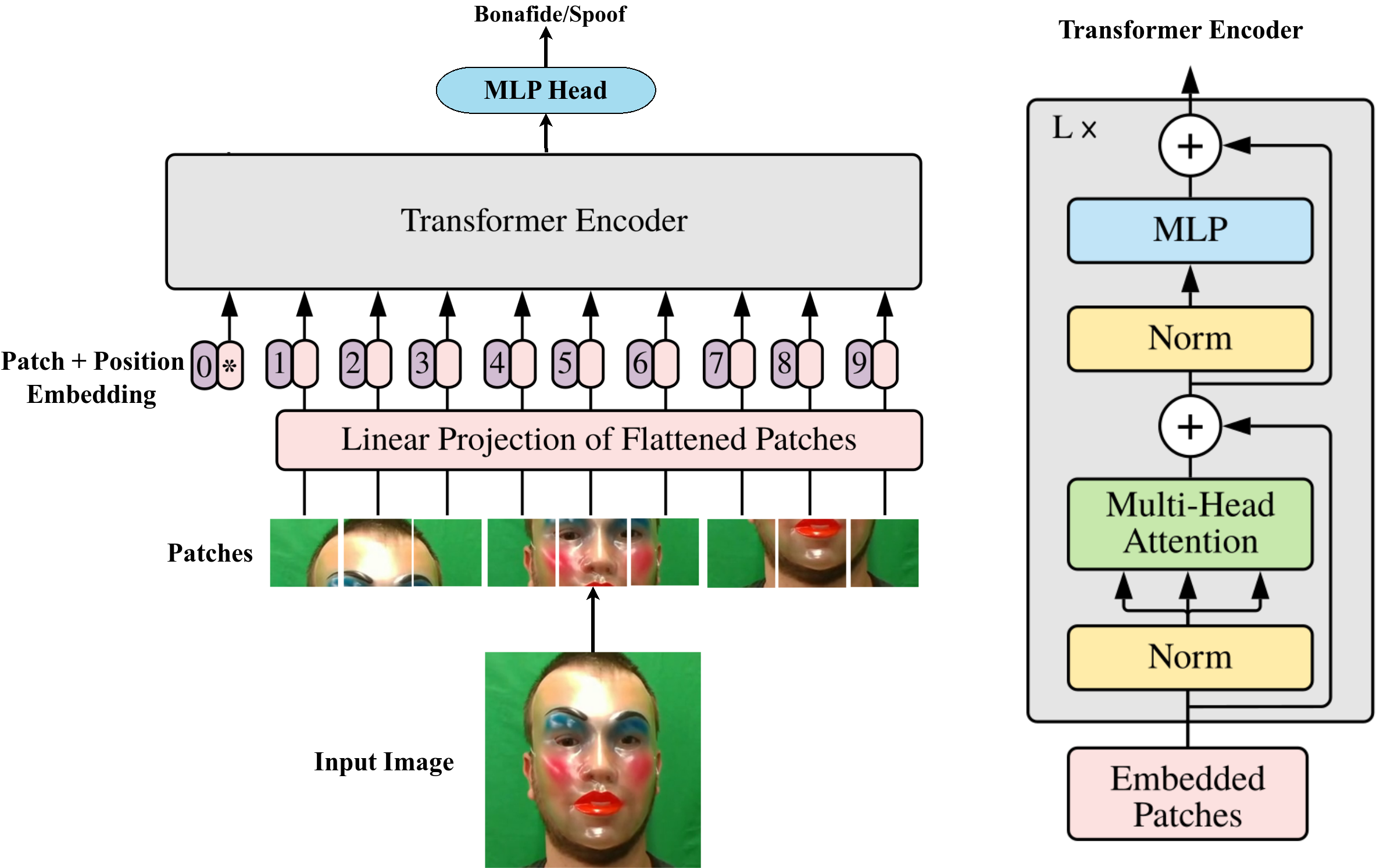}
        \caption{Vision Transformer model adapted for the presentation attack detection (PAD) task. The final layer is replaced and fine-tuned for the binary classification task (Image adapted from \cite{dosovitskiy2020image} and \cite{vaswani2017attention}).}
\label{fig:framework}
\vspace{5pt}
\end{figure}

In most cases, the amount of data available to train a PAD model is very limited. This limits the possibility of training deep architectures from scratch. And from the literature, transfer learning from a pre-trained network has proven to be an effective strategy to deal with the limited data problem. Moreover, a pre-trained network could also help with addressing the domain shift as the model has seen a wide variety of images in different environments.  

In this work, we investigate the effectiveness of the vision transformer model \cite{dosovitskiy2020image} for the zero-shot presentation attack detection problem. We compare the proposed method with both state-of-the-art methods and fine-tuned CNN models. Specifically, we investigate the performance of this method in challenging unseen attack and cross-database scenarios.

To the best knowledge of the authors, this is the first work using vision transformers for the presentation attack detection task. The main contributions of this work are listed below.

\begin{itemize}
\setlength{\itemsep}{1pt}
\setlength{\parskip}{0pt}
\setlength{\parsep}{0pt}

\item Introduces a simple yet effective Vision Transformer-based PAD framework. 
\item Shows the effectiveness of the vision transformer framework in an unrelated downstream task while adapting only a minimal number of parameters in the training stage.  
\item The proposed approach has been extensively evaluated in challenging unseen and cross-database conditions and it achieves the state of the art performance, outperforming other baselines by a large margin.
\end{itemize}

Additionally, the source code and protocols to reproduce the results are available publicly\footnote{Source code: \url{https://gitlab.idiap.ch/bob/bob.paper.ijcb2021_vision_transformer_pad}} to allow further extension of the work.

\section{Related work}

\paragraph{Presentation Attack Detection:} The majority of face presentation attack detection methods deal with the detection of 2D attacks. And most of these methods capitalize on capturing the quality degradation of the samples during recapture. Feature-based methods such as as motion patterns \cite{anjos2011counter}, Local Binary Patterns (LBP) \cite{boulkenafet2015face}, image quality features \cite{galbally2014image}, and image distortion analysis \cite{wen2015face} have been utilized over the years for PAD. Most of the recent PAD methods use CNN based approaches \cite{liu2018learning,george-icb-2019, atoum2017face}. Since many of these methods depend on quality degradation during recapture to distinguish attacks from bonafide, they may not be suitable for the detection of sophisticated attacks like 3D masks and partial attacks. Since the quality of attack instruments improves over time, the PAD methods used should be robust against unseen attacks as well. 

Multi-channel methods have been proposed in the literature as a solution to handle a wide variety of attacks \cite{raghavendra2017extended,steiner2016reliable,raghavendra2017extended,erdogmus2014spoofing,Bhattacharjee:256262}. The core idea of multi-channel/spectral methods is the usage of complementary information from different channels making it extremely hard for attackers to fool the PAD systems. An attacker would have to replicate the properties of a bonafide sample across different sensing domains, making it difficult depending on the channels used in the PAD system. In \cite{george_mccnn_tifs2019}, George \textit{et al}. presented a multi-channel face presentation attack detection framework using color, depth, infrared and thermal channels. Several recent works have also achieved good PAD performance utilizing multi-channel information \cite{heusch2020deep,george_mcocm_tifs2020,George_Idiap-RR-12-2020}.

Though multi-channel information could alleviate the issues with the PAD systems, the cost of the hardware increases with additional channels used. Moreover, it limits the widespread adoption of the PAD system, leaving legacy face recognition systems unprotected. An RGB-only PAD framework, which performs robustly across unseen attacks and environments is essential for the secure usage of legacy face recognition systems. 

\vspace{-10pt}
\paragraph{Transformer Models:} The Transformer models proposed in \cite{vaswani2017attention} introduced a novel approach towards sequence transduction tasks obviating the need for convolution or recurrence mechanisms. Transformers essentially capitalize on the attention mechanism to model dependencies between input and output. As opposed to other recurrent methods, Transformer allows significant parallelization in sequence tasks, while achieving the state of the art results in many tasks. There have been several attempts to use the Transformer framework for vision tasks, some of the methods used initial layers of pre-trained models \cite{liu2020endtoend} to get features to be used in the Transformer model. Various forms of position encodings, i.e., fixed or learned are also added to the features before passing to the Transformer layers. The work in \cite{carion2020end} used the encoder-decoder structure of the Transformer for the object detection task.

\section{Proposed method}

In this work, we propose to use transfer learning from a pre-trained Vision Transformer model for zero-shot face anti-spoofing task. The different stages of the framework are detailed in the following sections.

\subsection{Preprocessing}
\label{subsec:preprocess}
The CNN model accepts images of resolution $224 \times 224$. To avoid the contribution from the background and other database biases, we crop the face regions in the preprocessing stage. First, face detection and landmark localization are performed using the MTCNN \cite{zhang2016joint} algorithm. The detected faces are aligned so that the eye centers are horizontally aligned. After this alignment, the images are cropped to a resolution of $224 \times 224$.
\subsection{Network Architecture}
The proposed framework uses the recently proposed vision transformer \cite{dosovitskiy2020image} architecture as its back-bone. The details are given in the following sections.
\vspace{-10pt}

\paragraph{The Vision Transformer model:} Transformers were initially proposed by Vaswani \textit{et al.} \cite{vaswani2017attention} for machine translation applications. These models leverage the attention mechanism and this framework has found applications in many natural language, audio, and vision tasks. The attention layers \cite{bahdanau2014neural} aggregate information from the entire length of the input sequence. Transformers introduced self-attention layers that scan through and update each element in a sequence using the information from the whole sequence. Essentially, they explicitly model all the pairwise interactions between the components in the input sequence. Recently authors in \cite{dosovitskiy2020image} applied the standard transformer with minimal modifications for the image classification task. An image is divided into patches, and embeddings obtained from the patches are used as the sequence input for the transformer. The vision transformers introduce a new way for image classification instead of using convolutional layers. A sequence of image patches is used as the input followed by transformer layers.

While trained with large amounts of data, the vision transformer models outperform the state-of-the-art methods in many vision benchmarks. However, retraining such a large model from scratch is very computationally expensive. However, fine-tuning offers a way to utilize these powerful models in limited data scenarios without requiring much computational power. We used the model trained with $16 \times 16$ patches, meaning the input sequence length will be the number of patches $\frac{HW}{16^2}$, where $H$ and $W$ are the height and width of the input image in pixels. In addition to patch embeddings, a 1D positional embedding is also added to retain positional information. After the transformer layers, an MLP head consisting of a fully connected layer was added for the classification task. 

In this work, we investigate the transferability of the pre-trained Vision Transformer model for the PAD task. Specifically, we replace the last layer with a fully connected layer with one output node and we fine-tune the model using binary cross-entropy loss (BCE). We have conducted experiments with adapting a different set of layers to find the effect of fine-tuning when trained with a small dataset. The framework used for PAD is depicted in Fig. \ref{fig:framework}.
\vspace{-10pt}

\paragraph{Implementation details:} We adapted the Vision Transformer base network described in \cite{dosovitskiy2020image}, from the open-source implementation provided in \cite{timm2020}. Pretrained weights provided was used to initialize the network. Specifically, we used the ``base'' variant of the pre-trained model, made available for an image size of $224 \times 224$, with a patch size of $16 \times 16$. We used the model with the same resolution as other baselines ($224 \times 224$), to enable ready comparison between other ImageNet pre-trained models. Data augmentation was performed during the training phase with random horizontal flips with a probability of 0.5. The network was supervised with binary cross-entropy loss (BCE), with a fixed learning rate of $1\times10^{-4}$ and a weight decay parameter of  $1\times10^{-5}$. A batch size of $16$ was used (due to the large size of the model, and memory constraints) during training. We used the standard Adam Optimizer \cite{kingma2014adam}, for training the model on a GPU grid for 20 epochs. The best model was selected based on the minimum loss in the validation set. The architecture was implemented using the PyTorch \cite{paszke2017automatic} library, and the training and evaluation components were implemented using the \textit{Bob\footnote{\url{https://www.idiap.ch/software/bob/}}} \cite{bob2017} library to make it easy to reproduce the results.

\section{Experiments}
Details of the databases used and the experimental results with the proposed approach are detailed in this section.

\subsection{Databases}  

A wide variety of attacks are required to evaluate the performance of algorithms against unseen attacks. Most of the publicly available PAD datasets are limited to 2D print and replay attacks. Hence, we selected two publicly available datasets that contain a wide variety of 2D, 3D, and partial attacks, namely \hqwmca and \siwm datasets.



\textbf{\hqwmca dataset}: The High-Quality Wide Multi-Channel Attack (\hqwmca) dataset \cite{heusch2020deep,Mostaani_Idiap-RR-22-2020} consists of 2904 short multi-modal video recordings of both bonafide and presentation attacks. The database includes both obfuscation and impersonation attacks and the attack categories present are print, replay, rigid mask, paper mask, flexible mask, mannequin, glasses, makeup, tattoo, and wig. The number of bonafide subjects available is 51 and the dataset contains several data streams captured synchronously such as color, depth, thermal, infrared (spectra), and short-wave infrared (spectra). In this work, we utilize only the RGB data stream from the dataset. The RGB videos are captured at a resolution of $1920 \times 1200$.

We have created leave-one-out (LOO) attack protocols in \hqwmca by leaving out one attack type in the train and development set. The evaluation set consists of bonafide and the attack type which was left out. These sub-protocols constitute the zero-shot (or leave-one-out) protocols, which emulate the scenario of encountering an unseen attack type in a real-world scenario. We have also used the \textit{grandtest} protocol which consists of attacks distributed in the train, development, and test sets (with disjoint identities across folds), specifically for cross-database performance evaluation.

\textbf{\siwm dataset}: The Spoof in the Wild database with Multiple Attack Types (\siwm ) \cite{Liu_2019_CVPR} again consists of a wide variety of attacks captured using an RGB camera. The number of subjects present is 493, with 660 \textit{bonafide} and 968 attack samples with a total of 1628 files. There are 13 different sub-categories of attacks,  collected in different sessions, pose, lighting, and expression (PIE) variations. The attacks consist of various types of masks, makeups, partial attacks, and 2D attacks. The RGB videos are available in $1920 \times 1080$ resolution \footnote{The \siwm dataset is currently not publicly available due to a possible revision of the dataset. However, we have performed the experiments before the retraction of the dataset, and have obtained permission from the authors of the dataset to include the results in the current manuscript. The results correspond to the original release of \siwm as used in \cite{Liu_2019_CVPR}.}.

We use the leave-one-out (LOO) testing protocols available with the \siwm \cite{Liu_2019_CVPR} dataset for our experiments. The protocols available with the dataset consists of only \textit{train} and \textit{eval} sets. In each of the LOO protocols, the training set consists of 80\% percentage of the bonafide data and 12 types of spoof samples. The test set consists of 20\% of \textit{bonafide} data and the attack which was left out in the training set.
We created a subset of the train set (5\%), as the $dev$ set for model selection. In addition to the protocols available with the dataset, a \textit{grandtest} protocol was also created (as done in \cite{george_mcocm_tifs2020}) specifically for cross-database testing with attacks distributed more or less equally across folds.

\subsection{Metrics} 

For the evaluations in \hqwmca dataset, we have used the ISO/IEC 30107-3 metrics \cite{ISO}, Attack Presentation Classification Error Rate (APCER), and Bonafide Presentation Classification Error Rate (BPCER) along with the Average Classification Error Rate (ACER) in the $eval$ set. We compute the threshold in the $dev$ set for a BPCER value of 1\%. The ACER in the $eval$ set is calculated as the average of APCER and BPCER computed at this threshold.

For the \siwm database, to enable comparison with other state-of-the-art methods, we follow the same method of reporting results as compared to \cite{Liu_2019_CVPR}. Specifically, we apply a predefined threshold on the $eval$ set of all the protocols. The ACER, APCER, and BPCER are computed using a fixed threshold of 0.5 on all the sub-protocols. Additionally, the equal error rate (EER) is also reported in the $eval$ set.

For cross-database testing, Half Total Error Rate (HTER) is used following the convention in \cite{george_mcocm_tifs2020}, which computes the average of False Rejection Rate (FRR) and the False Acceptance Rate (FAR).

HTER is computed in the $eval$ set using the threshold computed in the $dev$ set using the EER criterion.  

\subsection{Baseline methods}

We have implemented three CNN-based baselines to compare with the proposed method. Since the proposed method
is based on transfer learning, we compare the proposed method with transfer learning from two popular architectures
for image classification namely \textit{ResNet} and \textit{DenseNet} architectures. Besides, we have implemented 
\textit{DeepPixBiS} architecture from literature which was specifically designed for presentation attack detection task. In addition to the implemented baselines, we compare the proposed approach with state-of-the-art methods from the literature in the \siwm dataset. The details of the baseline methods implemented are given below.\\
\textbf{\textit{ResNetPAD}} : Here we take the standard pre-trained \textit{ResNet} model \cite{he2016deep}, specifically we used the \textit{ResNet101} variant of the architecture. We replace the final layer with a new fully-connected layer making it suitable for binary classification. And while training, only the final fully connected layer is adapted.\\
\textbf{\textit{DenseNetPAD}}: Similarly, here we take the standard \textit{DenseNet} model \cite{huang2017densely} for fine-tuning. We used the \textit{DenseNet161} variant of the architecture in our experiments. Here again, we replace the final layer with a new fully-connected layer for binary classification and during training, only the last layer is adapted.\\
\textbf{\textit{DeepPixBiS}}:  This is a CNN based system \cite{george-icb-2019} which achieved good intra as well as cross-database performance in challenging OULU-NPU \cite{boulkenafet2017oulu} dataset. The network was trained using both binary and pixel-wise binary loss functions. The usage of pixel-wise loss acts as an auxiliary loss function forcing the network to learn a robust classifier.\\
\textbf{\textit{\textbf{ViTranZFAS}}}: This is our final proposed framework. Essentially, we take the pre-trained vision transformer model \cite{dosovitskiy2020image} and remove the final classification head. A new fully connected layer is added on top of the embedding followed by a sigmoid layer. The network is then trained using binary cross-entropy loss function, adapting only the final fully connected layer during training. 

\section{Experiments} 
\begin{table*}[ht!]
\centering
\caption{Performance of the baseline systems and the proposed method in \textbf{unseen} protocols of \hqwmca dataset. The values reported are ACER's obtained in the $eval$ set with a threshold computed for BPCER 1\% in $dev$ set.}
\label{tab:unseen_hqwmca}

\resizebox{0.9\textwidth}{!}
{
\begin{tabular}{l|r|r|r|r|r|r|r|r|>{\columncolor[gray]{0.8}}r}
\toprule
Method &  Flexiblemask &  Glasses &  Makeup &  Mannequin &  Papermask &  Rigidmask &  Tattoo &  Replay &  Mean $\pm$Std \\
\midrule
DenseNetPAD                                       &         28.20 &    45.50 &   36.60 &       0.40 &       6.90 &      12.70 &    4.60 &   32.40 & 20.91$\pm$15.76 \\
ResNetPAD                                      &         39.10 &    42.00 &   41.20 &       2.80 &       0.50 &      21.30 &   28.60 &   21.50 & 24.62$\pm$15.33 \\
DeepPixBiS \cite{george-icb-2019}                                        &          5.80 &    49.30 &   23.80 &       0.00 &       0.00 &      25.90 &   13.60 &    6.00 & 15.55$\pm$15.76 \\ \hline
\textbf{\textit{ViTranZFAS (FC)}}                         &          2.60 &    15.90 &   25.80 &       2.70 &       2.30 &       9.50 &    2.40 &   12.40 &  \textbf{9.20$\pm$ 7.99 } \\
\bottomrule
\end{tabular}
}
\end{table*}

We have conducted an extensive set of experiments in both intra and intra-dataset scenarios in both \hqwmca and \siwm datasets. Specifically, we evaluate the baselines and the proposed approach in unseen attack environments (zero-shot) protocols as it indicates the performance of these PAD systems encountering real-world attacks that were not seen during training time.
\vspace{-10pt}
\paragraph{Results in \textbf{\textit{HQ-WMCA}} dataset:} We have performed experiments using all the LOO protocols in the \hqwmca dataset and the results are tabulated in Table \ref{tab:unseen_hqwmca}. The values reported are the ACER in the $eval$ set corresponding to a threshold found from the $dev$ set (using  BPCER 1\% criterion). It can be seen that the proposed achieves much better performance than the baseline methods, achieving an average ACER of $9.020 \pm 7.99$ \%. This result is very promising since only the last fully connected layer was retrained for classification. 
\vspace{-10pt} 
\paragraph{Results in \textbf{\textit{SiW-M}} dataset:} The \siwm dataset contains a wide variety of attacks. We have performed experiments with the zero-shot protocols and the
results are tabulated in Table \ref{tab:siwm_results}. In this database, the proposed approach achieves a large performance improvement, nearly half of the error rate as compared to the state-of-the-art methods. The proposed approach performs well on most of the sub-protocols and achieves a mean EER or $6.72 \pm 5.66$ \%. 

\begin{table*}[ht!]
\small
  \centering
  \caption{Performance of the proposed framework in the leave one out protocols in \siwm dataset.}
  \label{tab:siwm_results}
  \resizebox{\textwidth}{!}
{
{

  \begin{tabular}{l|l|c|c|c|c|c|c|c|c|c|c|c|c|c|>{\columncolor[gray]{0.8}}r}
  \toprule
  \multirow{2}{*}{Methods} & \multirow{2}{*}{Metrics (\%)} &\multirow{2}{*}{Replay}& \multirow{2}{*}{Print} & \multicolumn{5}{c|}{Mask Attacks}  & \multicolumn{3}{c|}{Makeup Attacks}  &  \multicolumn{3}{c|}{Partial Attacks} & \multirow{1}{*}{Average}\\ \cline{5-15}
   &&&  & Half & Silicone & Trans. & Paper & Manne. & Obfusc. & Imperson. & Cosmetic & Funny Eye & Paper Glasses & Partial Paper &\\ \midrule

    \multirow{4}{*}{Auxiliary \cite{liu2018learning} }
     & APCER &$23.7$& $7.3$ & $27.7$ & $18.2$ & $97.8$ & $8.3$ & $16.2$ & $100.0$ & $18.0$ & $16.3$ & $91.8$ & $72.2$ & $0.4$ & $38.3\pm37.4$  \\ \cline{3-16}
     & BPCER &$10.1$& $6.5$ & $10.9$ & $11.6$ & $6.2$ & $7.8$ & $9.3$ & $11.6$ & $9.3$ & $7.1$ & $6.2$ & $8.8$ & $10.3$ & $8.9\pm2.0$  \\ \cline{3-16}
     & ACER &$16.8$& $6.9$ & $19.3$ & $14.9$ & $52.1$ & $8.0$ & $12.8$ & $55.8$ & $13.7$ & $11.7$ & $49.0$ & $40.5$ & $5.3$ &$23.6\pm18.5$  \\ \cline{3-16}
     & EER &$14.0$& $4.3$ & $11.6$ & $12.4$ & $24.6$ & $7.8$ & $10.0$ & $72.3$ & $10.1$ & $9.4$ & $21.4$ & $18.6$ & $4.0$  & $17.0\pm17.7$ \\ \midrule

    \multirow{4}{*}{Deep Tree Network \cite{Liu_2019_CVPR}}

     & APCER &$1.0$& $0.0$ & $0.7$ & $24.5$ & $58.6$ & $0.5$ & $3.8$ & $73.2$ & $13.2$ & $12.4$ & $17.0$ & $17.0$ & $0.2$ & $17.1\pm23.3$  \\ \cline{3-16}
     & BPCER &$18.6$& $11.9$ & $29.3$ & $12.8$ & $13.4$ & $8.5$ & $23.0$ & $11.5$ & $9.6$ & $16.0$ & $21.5$ & $22.6$ & $16.8$ & $16.6\pm6.2$  \\ \cline{3-16}
     & ACER &$9.8$& $6.0$ & $15.0$ & $18.7$ & $36.0$ & $4.5$ & $7.7$ & $48.1$ & $11.4$ & $14.2$ & $19.3$ & $19.8$ & $8.5$ & $16.8\pm11.1$  \\ \cline{3-16}
     & EER &$10.0$& $2.1$ & $14.4$ & $18.6$ & $26.5$ & $5.7$ & $9.6$ & $50.2$ & $10.1$ & $13.2$ & $19.8$ & $20.5$ & $8.8$ & $16.1\pm12.2$   \\ \midrule

  \multirow{4}{*}{MCCNN (BCE+OCCL)-GMM \cite{george_mcocm_tifs2020}}
  &APCER &  11.79 &  9.53 &   3.12 &   3.70 &  39.20 &    0.00 &   3.12 &   44.57 &   0.00 &  21.60 &  19.34 &  35.55 &   0.00 & $ 14.7 \pm 15.9 $ \\ \cline{3-16}
  &BPCER &  13.44 & 16.15 &  16.26 &  20.23 &  11.11 &   13.74 &   8.66 &   15.23 &  12.67 &  10.42 &  14.31 &  18.40 &  27.33 & $ 15.2 \pm  4.8 $ \\ \cline{3-16}
  &ACER  &  12.61 & 12.84 &   9.69 &  11.97 &  25.16 &    6.87 &   5.89 &   29.90 &   6.34 &  16.01 &  16.83 &  26.97 &  13.66 & $ 14.9 \pm  7.8 $ \\ \cline{3-16}
  &EER   &  12.82 & 12.94 &  11.33 &  13.70 &  13.47 &    0.56 &   5.60 &   22.17 &   0.59 &  15.14 &  14.40 &  23.93 &   9.82 & $ 12.0 \pm  6.9 $ \\ \midrule

  \multirow{4}{*}{DenseNetPAD}
  &APCER &   28.32 &  11.00 &     26.81 &         23.84 &            39.88 &       0.04 &           3.22 &        70.69 &           0.01 &     38.21 &            72.66 &         53.37 &      3.46 & 28.58 $\pm$24.48 \\ \cline{3-16}
  &BPCER &   12.48 &  12.44 &     12.63 &         13.63 &            12.31 &      13.18 &          13.81 &        11.50 &          13.93 &     11.60 &            12.59 &         12.00 &     12.48 & 12.66 $\pm$ 0.75 \\ \cline{3-16}
  &ACER  &   20.40 &  11.72 &     19.72 &         18.74 &            26.10 &       6.61 &           8.51 &        41.10 &           6.97 &     24.90 &            42.63 &         32.69 &      7.97 & 20.62 $\pm$12.00 \\ \cline{3-16}
  &EER   &   16.49 &  12.10 &     15.43 &         17.03 &            19.91 &       5.58 &          10.44 &        24.21 &           3.67 &     17.66 &            29.12 &         22.26 &      9.78 & 15.67 $\pm$ 7.03 \\ \midrule

  \multirow{4}{*}{ResNetPAD}
  &APCER &   32.08 &  18.88 &     34.64 &         21.22 &            33.96 &       0.00 &           2.78 &        94.38 &           0.00 &     35.38 &            72.92 &         18.83 &      2.72 & 28.29 $\pm$ 27.22 \\ \cline{3-16}
  &BPCER &    9.47 &   9.48 &     10.51 &         10.95 &            10.50 &      11.33 &          10.79 &         9.39 &          11.03 &     10.17 &            10.79 &         11.09 &     10.68 & 10.48 $\pm$  0.63 \\ \cline{3-16}
  &ACER  &   20.78 &  14.18 &     22.57 &         16.09 &            22.23 &       5.66 &           6.79 &        51.88 &           5.51 &     22.77 &            41.86 &         14.96 &      6.70 & 19.38 $\pm$ 13.45 \\ \cline{3-16}
  &EER   &   15.17 &  11.81 &     18.25 &         14.69 &            16.87 &       1.96 &           7.95 &        33.27 &           4.19 &     17.72 &            28.85 &         12.86 &      7.86 & 14.73 $\pm$  8.54 \\ \midrule

    \multirow{4}{*}{DeepPixBiS \cite{george-icb-2019}}

    & APCER & 19.18 & 8.97 &  1.74 &   21.30 &   60.68 &   0.00 &  1.00 &    100.00 &     0.00 &    26.90 &    64.66 &   77.52 &  0.29 & $ 29.4 \pm $ 34.4 \\ \cline{3-16}
  & BPCER &  8.70 & 7.63 & 11.03 &   11.76 &   10.27 &   8.85 &  8.63 &     10.53 &    11.60 &    10.99 &    10.31 &   10.23 &  7.10 & $  9.8 \pm $  1.4 \\ \cline{3-16}
  & ACER  & 13.94 & 8.30 &  6.38 &   16.53 &   35.47 &   4.43 &  4.81 &     55.27 &     5.80 &    18.95 &    37.48 &   43.87 &  3.69 & $ 19.6 \pm $ 17.4 \\ \cline{3-16}
  & EER   & 11.68 & 7.94 &  7.22 &   15.04 &   21.30 &   3.78 &  4.52 &     26.49 &     1.23 &    14.89 &    23.28 &   18.90 &  4.82 & $ 12.3 \pm $  8.2 \\ \midrule

  \multirow{4}{*}{\textbf{\textit{ViTranZFAS (FC)}}}
  &APCER &  38.27 &  5.81 &   5.00 &   4.62 &   5.47 &    0.00 &   0.32 &   12.55 &   0.00 &  18.32 &  61.81 &  0.29 &  0.13   & $11.74 \pm 17.75 $ \\ \cline{3-16}
  &BPCER &   4.82 &  5.87 &   6.27 &   5.52 &   6.33 &    5.68 &   6.02 &    6.22 &   6.66 &   5.62 &   5.46 &  7.03 &  5.50   & $ 5.92 \pm  0.56 $ \\\cline{3-16}
  &ACER  &  21.55 &  5.84 &   5.63 &   5.07 &   5.90 &    2.84 &   3.17 &    9.38 &   3.33 &  11.97 &  33.63 &  3.66 &  2.82   & $ 8.83 \pm  8.73 $ \\ \cline{3-16}
  &EER   &  15.20 &  5.84 &   5.80 &   4.99 &   5.95 &    0.12 &   3.25 &    9.89 &   0.46 &  10.76 &  20.19 &  2.96 &  1.97   & $ \textbf{6.72} \pm  \textbf{5.66} $ \\

    \bottomrule

  \end{tabular}
  }
  }
\end{table*}

\begin{table*}[ht!]
\centering
\caption{Performance of the Vision Transformer network when fine tuning different set of layers in \textbf{unseen} protocols of \hqwmca dataset. The values reported are ACER's obtained in $eval$ set with a threshold computed for BPCER 1\% in $dev$ set.}
\label{tab:unseen_hqwmca_ablation}

\resizebox{0.9\textwidth}{!}
{

\begin{tabular}{l|r|r|r|r|r|r|r|r|>{\columncolor[gray]{0.8}}r}
\toprule
Adapted Layers &  Flexiblemask &  Glasses &  Makeup &  Mannequin &  Papermask &  Rigidmask &  Tattoo &  Replay &  Mean $\pm$Std \\
\midrule

\textit{ViTranZFAS (ALL)}                           &          2.40 &    44.90 &   21.10 &       0.00 &       0.20 &      21.60 &    0.60 &   25.90 & 14.59$\pm$15.42 \\
\textit{ViTranZFAS (E+FC)}                   &          5.50 &    47.00 &   23.40 &       1.90 &      19.40 &      11.60 &    2.10 &   11.50 & 15.30$\pm$13.99 \\
\textbf{\textit{ViTranZFAS (FC)}}                         &          2.60 &    15.90 &   25.80 &       2.70 &       2.30 &       9.50 &    2.40 &   12.40 &  \textbf{9.20$\pm$ 7.99 } \\

\bottomrule
\end{tabular}
}
\end{table*}
\vspace{-10pt}

\paragraph{Analysis of training strategies:} It was observed that fine-tuning the last layer alone was achieving state-of-the-art performance. Here we examine the effectiveness of retraining different sets of layers in the \hqwmca dataset. We have considered three different settings for this study they are, 
\begin{itemize}
\setlength{\itemsep}{1pt}
\setlength{\parskip}{0pt}
\setlength{\parsep}{0pt}
\item \textit{ViTranZFAS (FC)}: Here, only the last fully connected layer is retrained, all the other layers are frozen, this corresponds to the fine-tuning scenario from a pre-trained model.
\item \textit{ViTranZFAS (ALL)} layers: All the layers from the architecture are adapted during training.
\item \textit{ViTranZFAS (E+FC)}: Here both the first embedding layers as well as the final fully connected layer are adapted.
\end{itemize}

The results for this set of experiments are shown in Table \ref{tab:unseen_hqwmca_ablation}. Clearly, adapting only the FC layer achieves the best results. Given the large amount of data used for training the pre-trained model, adapting other layers appears to be prohibitive with a limited amount of training data. 

\begin{figure*}[h!]
\centering
  \subfloat[\hqwmca]{\includegraphics[width=0.33\textwidth]{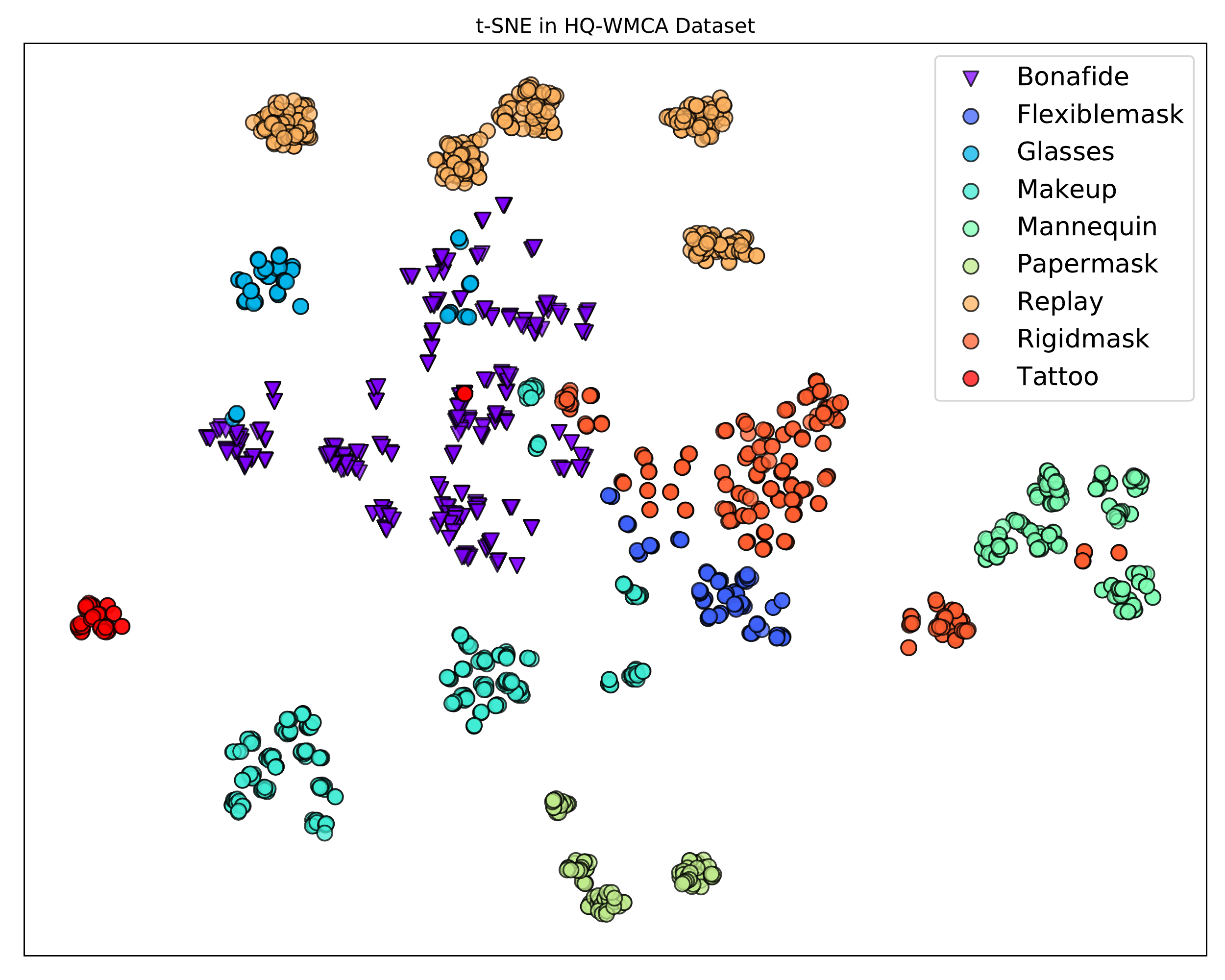}}%
\hfil
  \subfloat[\siwm]{\includegraphics[width=0.33\textwidth]{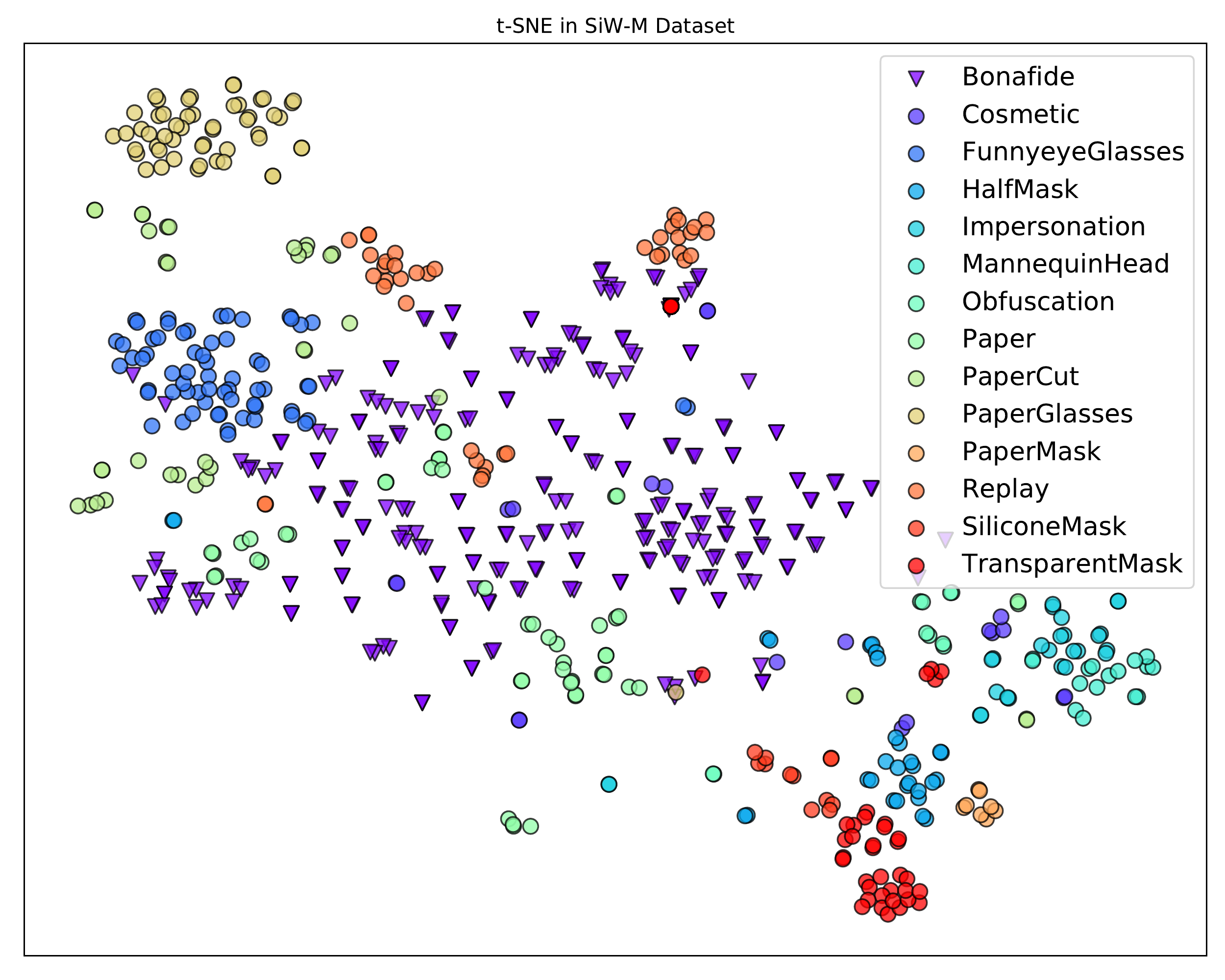}}%
  \hfil
  \subfloat[\hqwmca - \siwm]{\includegraphics[width=0.33\textwidth]{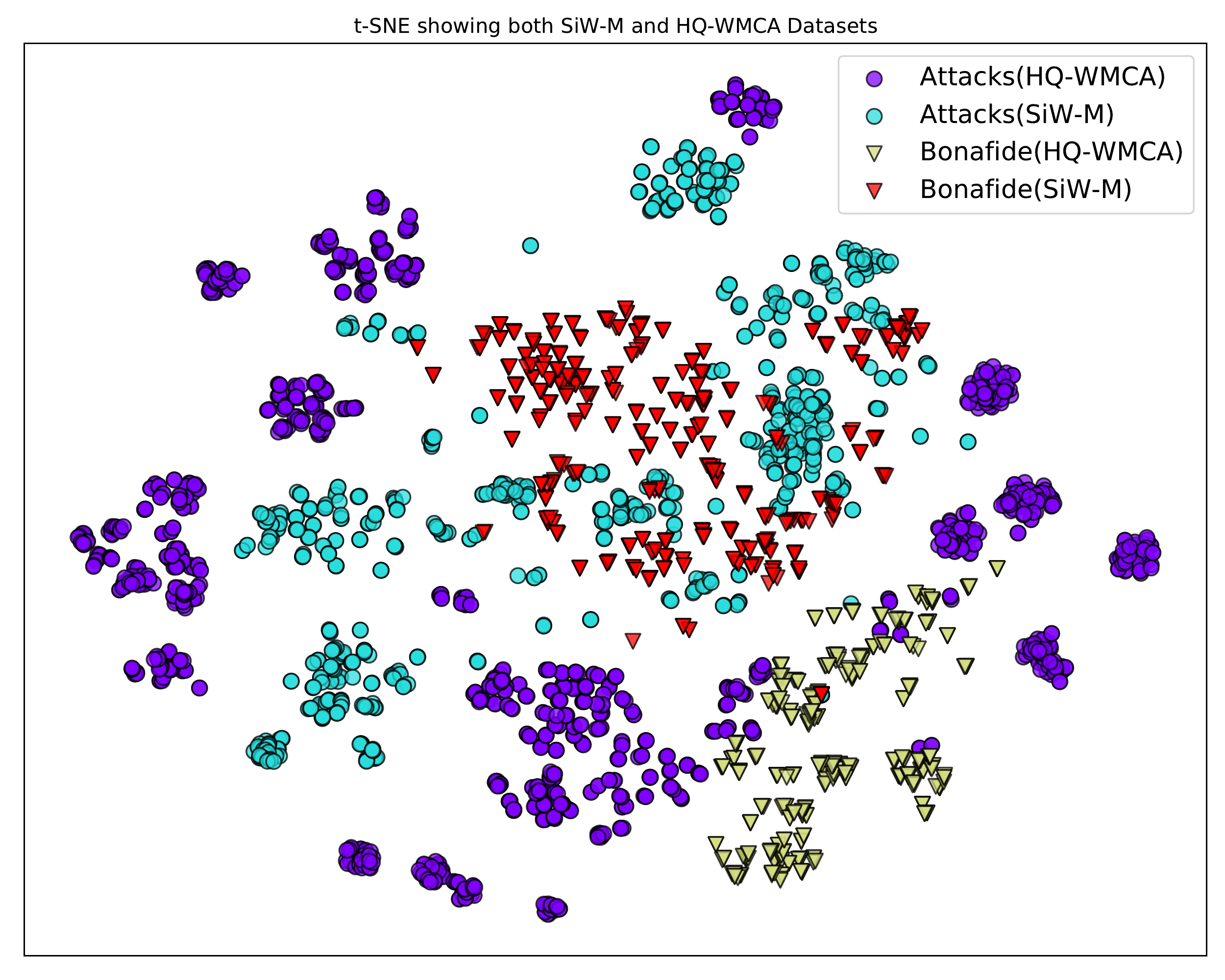}}%
\caption{t-SNE plots corresponding to the 768-dimensional feature from the Vision Transformer, the plot already shows some separability between bonafide and attacks in the features extracted from the pre-trained model.}
 \label{fig:tsne_plots}
\end{figure*}

Figure \ref{fig:tsne_plots} shows the t-SNE plots from the Vision Transformer embeddings (from the pre-trained models) on both \hqwmca and \siwm datasets individually and together (for the $eval$ set in the corresponding \textit{grandtest} protocols). It can be seen that there is already a good amount of separability between bonafide and the spoof samples in the feature space. This could explain the good performance of the proposed method just by adapting the final fully connected layer. 

\vspace{-10pt}

\begin{figure}[ht]
\centering
        \includegraphics[width=0.7\linewidth]{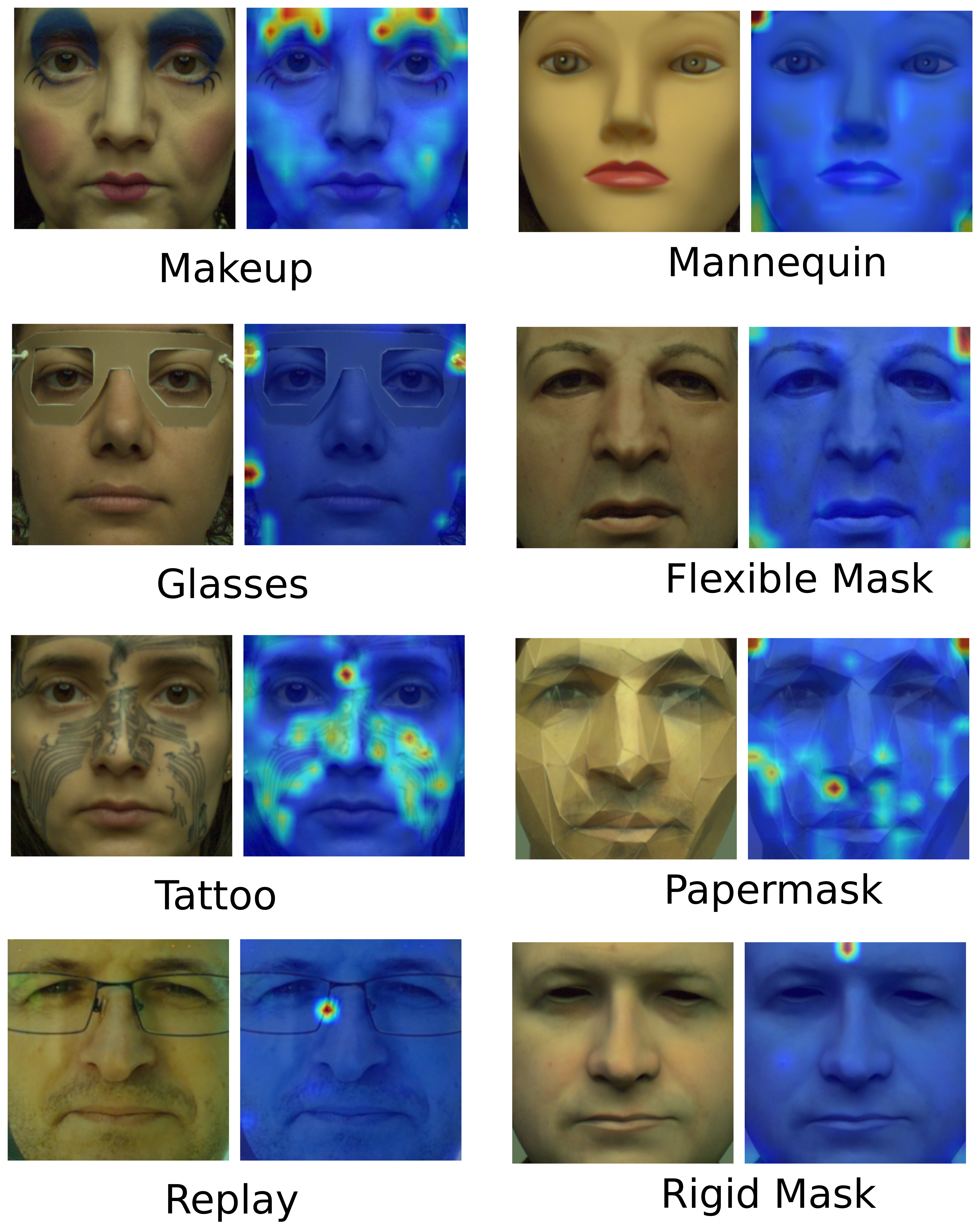}
        \caption{Relevancy maps for different types of attacks with \textit{ViTranZFAS (FC)} model in the grandtest protocol in \hqwmca dataset.}
\label{fig:lrp}
\end{figure}

\paragraph{Visualization of different classes:} To further understand the features contributing to the decisions, we computed the relevancy maps for different classes.
In \cite{chefer2020transformer}, the authors proposed a way to visualize the relevancy maps for Transformer networks. 
Essentially, their method assigns local relevance based on the deep Taylor decomposition which propagates the relevancy scores through the layers.
This method achieved the state of the art results compared to other methods, for computing the relevancy maps for Transformer networks. We have computed the relevancy maps for the fine-tuned Vision Transformer model (\textit{ViTranZFAS (FC)}). The relevance maps for various attack types are shown in Fig. \ref{fig:lrp}. 
Interestingly, in ``Makeup'' and ``Tattoo'' attacks, the network correctly identifies the regions of importance. For attacks like ``Replay'' and ``Rigid mask'' and the network struggles as the discriminative region is not localized, the spoof traces a spread out throughout the face in such attacks.

\vspace{-10pt}
\paragraph{Cross-database evaluations}: From the previous sections, it was clear that the proposed approach achieves much better performance as compared to the state-of-the-art methods in unseen attack scenarios. One main issue of PAD methods has been poor cross-database generalization, which is essential to ensure reliable performance in real-world deployment scenarios. To evaluate the generalization, we performed cross-database evaluations between \hqwmca and \siwm datasets. For each dataset, we trained the model using the \textit{grandtest} protocol of the corresponding dataset, and the resulting model is evaluated using the \textit{grandtest} protocol of the other dataset. The results are tabulated in table \ref{tab:cross_test_eer}. From the results, it can be seen that the proposed approach improves the cross-database performance by a large margin indicating the generalizability of the proposed approach. 

\begin{table}[h!]
\begin{center}
\footnotesize
\caption{The results from the cross-database testing between \siwm and \hqwmca datasets. HTER (\%) values computed in $eval$ set for threshold computed in $dev$ set using EER criteria are reported in the table.}
\label{tab:cross_test_eer}

\resizebox{0.99\columnwidth}{!}{%
\begin{tabular}{@{}l|c|>{\columncolor[gray]{0.8}}c|c|>{\columncolor[gray]{0.8}}c@{}}
\toprule
\multirow{3}{*}{Method} & \multicolumn{2}{c|}{\begin{tabular}[c]{@{}c@{}}trained on\\ \hqwmca \end{tabular}}                                     & \multicolumn{2}{c}{\begin{tabular}[c]{@{}c@{}}trained on\\ \siwm \end{tabular}}                                       \\ \cline{2-5}  
                        & \begin{tabular}[c]{@{}c@{}}tested on\\  \hqwmca \end{tabular} & \begin{tabular}[c]{@{}c@{}}tested on\\ \siwm \end{tabular} & \begin{tabular}[c]{@{}c@{}}tested on\\ \siwm \end{tabular} & \begin{tabular}[c]{@{}c@{}}tested on \\ \hqwmca \end{tabular} \\ \midrule


DenseNetPAD                   &11.40               &27.5             &10.4            &29.3            \\
ResNetPAD                     &13.50               &25.7             &10.4            &29.4            \\
DeepPixBiS \cite{george-icb-2019}                   &\textbf{4.60}                &25.6             &14.7            &38.1            \\ \hline
\textbf{\textit{ViTranZFAS (FC)}}        &5.60                &\textbf{14.7}             &\textbf{6.0}             &\textbf{12.7}            \\

 \bottomrule
\end{tabular}
}
\end{center}

\end{table}
\vspace{-10pt}
\paragraph{Computational Complexity:} Here we compare the parameters, and complexity of the baseline and the vision transformer model. The comparison is shown in Table \ref{tab:complexity}. It can be seen that the vision transformer model requires more computational and parameters as compared to the baselines. Though the network is complex, we just retrain just the last fully connected layer with just 768 neurons in our transfer learning setting making the training far easier. Distillation of the model \cite{tang2019distilling} to reduce the complexity could be a possible direction to address this limitation.

\begin{table}[h]
\begin{center}
\caption{Computational and parameter complexity comparison}
\label{tab:complexity}
  \resizebox{0.7\linewidth}{!}
{
\begin{tabular}{l|>{\columncolor[gray]{0.8}}r|>{\columncolor[gray]{0.8}}r}
\toprule
Model                 & Compute & Parameters \\ \midrule
ResNetPAD             & 7.85 GMac                & 42.5 M     \\
DenseNetPAD           & 7.82 GMac                & 26.47 M    \\
MCCNN(RGB) \cite{george_mcocm_tifs2020} & 10.88 GMac               & 37.73 M     \\
DeepPixBiS \cite{george-icb-2019}            & 4.64 GMac                & 3.2M       \\ \hline
\textbf{\textit{ViTranZFAS}}      & 16.85 GMac                & 85.8 M      \\ \bottomrule
\end{tabular}
}
\end{center}
\end{table}

\vspace{-10pt}
\paragraph{Discussions:} From the experimental results in both \hqwmca and \siwm datasets, it can be seen that the proposed method achieves state-of-the-art performance in challenging unseen attack scenarios. Surprisingly the proposed method achieves excellent cross-database generalization as well. Typical PAD models have a tendency to overfit to the nuances in specific datasets rather than focusing on the reliable spoof cues, resulting in poor generalization in cross-database scenarios. Using the pre-trained model provides a good prior, and training strategy adapting a minimal subset of layers reduces the chances of overfitting, achieving good performance in the challenging scenarios. The inherent properties of the vision transformers also boost the performance. Self-attention as opposed to convolutions helps to attend to all pairwise interactions in the lower layers itself. The large datasets used for pre-training the vision transformer models also improve the robustness. As shown in Fig. \ref{fig:lrp}, the model correctly focuses on the discriminative features. 

\section{Conclusions} 
\label{sec:conclusion}

In this work, we have shown the effectiveness of the vision transformer network for the zero-shot face anti-spoofing task. Essentially, just fine-tuning a pre-trained vision transformer model for the PAD task was sufficient to achieve the state-of-the-art performance in \hqwmca and \siwm datasets. In addition to excellent performance in unseen attacks, the proposed approach outperforms the state-of-the-art methods in cross-datasets evaluations by a large margin, indicating the efficacy of the proposed approach in generalizing to both unseen attacks and domains. The vision transformers could prove to be very beneficial in dealing with the current limitations of presentation attack detection systems. The datasets, source code, and protocols used are made available publicly to enable the further extension of the work.

To summarize, in this work we show that merely fine-tuning the last fully connected layer in vision transformers achieves state-of-the-art performance in both unseen attack and cross-database scenarios. Extensive evaluations show the effectiveness of the method. The superior performance in addressing two of the challenging issues (unseen attack and cross-database generalization) in the PAD task with minimal fine-tuning holds the potential to address the issues with PAD models. We hope that this work will motivate the biometrics community to investigate transformer models further.

\section*{Acknowledgment} 

Part of this research is based upon work supported by the Office of the
Director of National Intelligence (ODNI), Intelligence Advanced Research
Projects Activity (IARPA), via IARPA R\&D Contract No. 2017-17020200005.
The views and conclusions contained herein are those of the authors and
should not be interpreted as necessarily representing the official
policies or endorsements, either expressed or implied, of the ODNI,
IARPA, or the U.S. Government. The U.S. Government is authorized to
reproduce and distribute reprints for Governmental purposes
notwithstanding any copyright annotation thereon.

{\small
\bibliographystyle{ieee}
\bibliography{egbib}
}

\end{document}